\documentclass{article}

\usepackage[nonatbib, final]{nips_2016}

\usepackage[utf8]{inputenc} 
\usepackage[T1]{fontenc}    
\usepackage{lmodern}
\usepackage{hyperref}       
\usepackage{url}            
\usepackage{booktabs}       
\usepackage{amsfonts}       
\usepackage{nicefrac}       
\usepackage{microtype}      
\usepackage{graphicx}
\usepackage{algorithmicx}
\usepackage{amsmath}
\usepackage{algorithm}
\usepackage[noend]{algpseudocode}
\usepackage{standalone}
\usepackage{titling}
\usepackage[nonatbib, final]{nips_title}

\title{Multi-View Treelet Transform}

\author{
  Brian A. Mitchell \\
  Department of Computer Science\\
  University of California at Santa Barbara\\
  Santa Barbara, CA 93106 \\
  \texttt{brian\_a\_mitchell@cs.ucsb.edu} \\
  \And
  Linda R. Petzold \\
  Department of Computer Science \\
  Department of Mechanical Engineering \\
  University of California at Santa Barbara \\
  Santa Barbara, CA 93106 \\
  \texttt{petzold@cs.ucsb.edu} \\
}

\begin{document}


\maketitle

\begin{abstract}
    Current multi-view factorization methods make assumptions that are not acceptable for many kinds of data, and in particular, for graphical data with hierarchical structure.  At the same time, current hierarchical methods work only in the single-view setting.  We generalize the Treelet Transform to the Multi-View Treelet Transform (MVTT) to allow for the capture of hierarchical structure when multiple views are available.  Further, we show how this generalization is consistent with the existing theory and how it might be used in denoising empirical networks and in computing the shared response of functional brain data.  
\end{abstract}

\section{Introduction}

Much of the focus of modern machine learning and data analysis involves the collection and analysis of data from multiple perspectives, or views, of the same phenomenon.  In every application there is a common desire to better understand the mathematical structure of this phenomenon.  One approach to quantifying this structure is to jointly factorize the data matrices, where success is often measured by the total reconstruction error across all views:

\[ \textrm{Error} = \sum_{i=1}^M ||A_i - F_i|| + R. \]
\vspace{0mm} \\
Here, $||\cdot||$ is some matrix norm, $F_i$ is the reconstruction of $A_i$ after factorization, $R$ is a regularizer, and $M$ is the total number of views.  There are effectively two approaches to capturing consensus structure in multi-view factorization: construction of a common basis or construction of a common coefficient matrix.  The fundamental assumption for the former is that all views belong to the same space, whereas for the latter, the assumption is that the relative importance of the basis vectors is invariant across views.  \\
\vspace{0mm} \\
Though our method is not strictly a matrix factorization as we do not minimize an objective of the form given above, we compute a common basis across all views.  A frequently made assumption when computing such a basis is that the data matrices being factorized are low-rank \cite{Tang}.  When trying to capture structure in systems where such an assumption does not hold, these factorization methods do poorly.  Several methods \cite{MMF, DW, Treelet} have been proposed recently that allow for a different assumption to be made in computing a basis for some dataset, namely, that the data have a hierarchical structure.  We generalize one of these methods, the Treelet Transform, to the multi-view setting and provide a discussion on the theoretical issues raised by this generalization.  While our method can be applied to collections of asymmetric or non-square data matrices, we focus on its application to collections of symmetric matrices, namely, adjacency matrices of graphs.  We show how the Multi-View Treelet Transform (MVTT) outperforms low-rank methods and the single view Treelet Transform in capturing the hierarchical structure of a synthetic dataset.  Further, we give an example of how our method can reduce dependence on ad-hoc denoising methods.  Finally, we demonstrate its excellent performance on an fMRI shared response problem when compared with existing methods.  

\section{Related work}

The main contribution of this work is a generalization of the methods used to compute hierarchical bases, specifically, the Treelet Transform.  For this, we draw heavily from \cite{Treelet}.  As \cite{Treelet} is similar in its goals and approach to \cite{MMF, DW}, we have also drawn from these works.  Specifically, the authors in \cite{Treelet} have developed a method for computing a sparse, hierarchical basis.  We have found that the performance of these methods can be improved in the setting where multiple views of some phenomenon are available.  In generalizing the work of \cite{Treelet}, we compute a consensus representation across views in the form of a single basis.  Other works involve computing a common basis in the multi-view setting, but there are none that exploit the hierarchical structure of the data.  In \cite{Tang}, the data is assumed to have a low-rank structure; the algorithms proposed in \cite{BLM, PLS, GMA, CCA} all place constraints on the basis that can be computed, though none explicitly compute a single basis.  A number of methods exist that compute a coefficient matrix as their consensus representation.  Many of these methods are a form of Multi-View Non-Negative Matrix Factorization \cite{symNMF, multiNMF, equiNMF, multiFactorNMF}.  A similar approach is taken in computational neuroscience under the label of the Shared Response Model (SRM) \cite{SRM}.  In this model, a multi-view matrix factorization, assuming low-rank data, is computed with a consensus coefficient matrix.  

\section{Treelet Transform}

The Treelet Transform is presented in \cite{Treelet} as a sequence of local PCA's.  With each iteration of the algorithm, two similar columns are rotated to create a normalized ``sum'' column and a normalized ``difference'' column.  The difference column is excluded from further iterations, and as such, increasingly abstract, smooth representations are computed over the data with subsequent iterations.  The construction of these increasingly smooth representations gives rise to the interpretation of this algorithm as inducing a hierarchical clustering on the data.  The complete algorithm is shown in Algorithm 1.

\begin{figure}[h]
  \centering
  \includegraphics[width=0.6\linewidth]{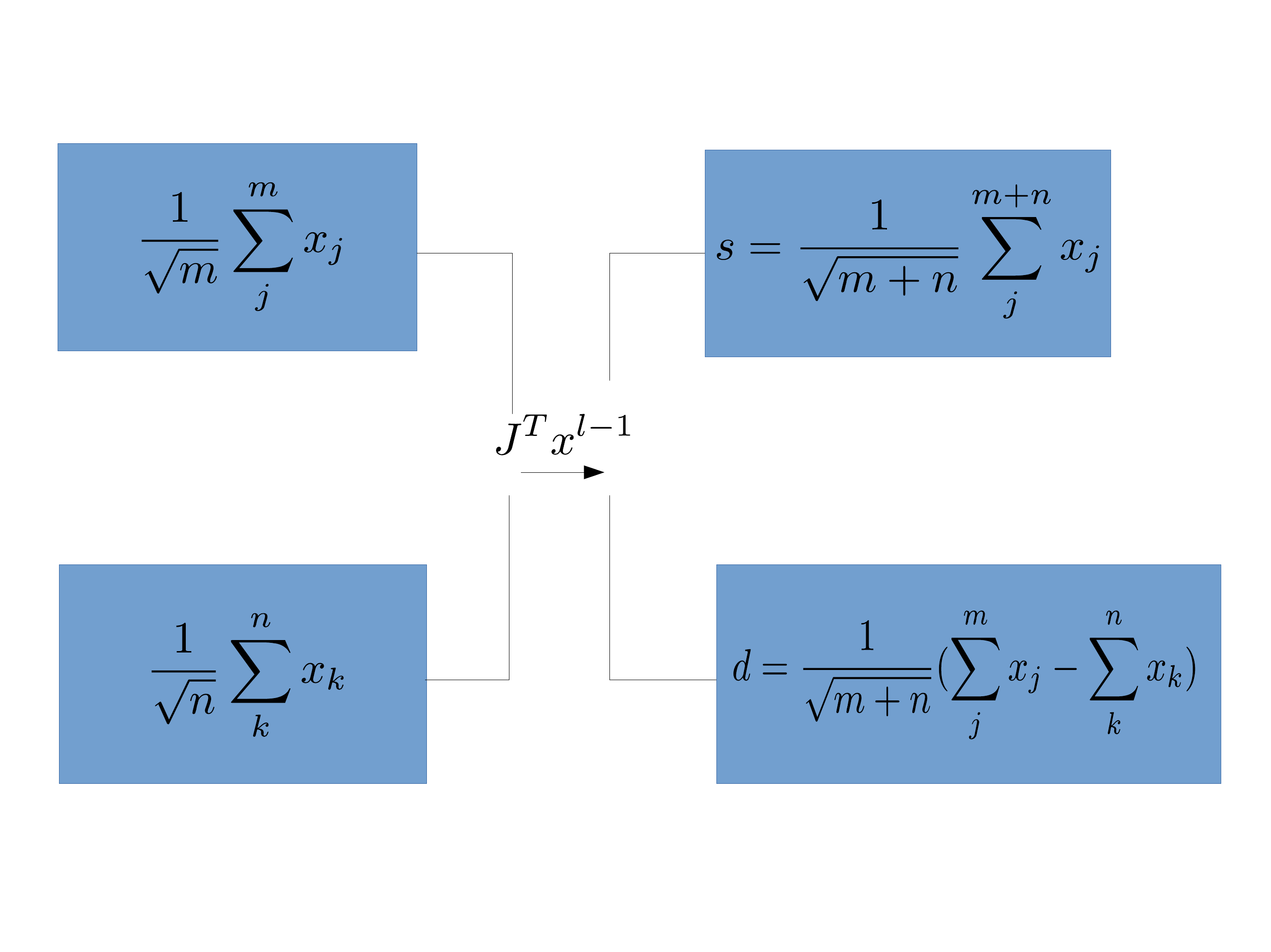}
  \caption{Shown is an example of one iteration of the Treelet Transform algorithm once two similar sum columns, $u$ and $v$ are found.  A rotation, $J$ is computed and then applied to the data matrix at iteration (i.e. hierarchical level) $l-1$ that rotates these two columns into $s$ (a new sum column) and $d$ (a new difference column).}
\end{figure}  

\begin{algorithm}[h]
\caption{}\label{euclid}
\begin{algorithmic}[1]
\Procedure{TreeletTransform}{$X$, $L$}
    \State Compute the covariance, $\Sigma^1$, and correlation coefficient matrices, $\rho^1$, from $X$
    \State $l \gets 1$, $S \gets \{1, ..., M\}$, $B^1 \gets \textrm{I}$
    \While{$l < L$}
      \State $(j,k) = \textrm{argmax}_{(j,k)} \rho[j,k]$
      \State Compute $J$ such that $J^T\Sigma^l J [j,k] = J^T\Sigma^l J [k,j] = 0$
      \State Drop column $j$ from $S$
      \State $B^{l+1} \gets B^l J$
      \State $\Sigma^{l+1} \gets J^T\Sigma^l J$
      \State $X^{l+1} \gets J^TX^l$
      \State $\rho^{l+1} \gets J^T\rho^l$
      \State $l++$
    \EndWhile
    \State return $B^L$
\EndProcedure
\end{algorithmic}
\end{algorithm}

Of note is the problem of selecting the parameter $L$, that is, the number of rotations to perform, which is equal to the height of the hierarchical tree.  In \cite{Treelet} it is suggested to use the best K-basis method, which raises the issue of selecting $K$.  We take a different approach: we propose a method for learning $K$ based on the data while leaving $L$ as a parameter.  This choice is based on the applications demonstrated in this paper, the data for which all have about the same hierarchical depth.  We fix $L=p/2$ for all experiments, based on the empirical performance of this value.  With regard to our method for learning $K$, we defer discussion of this issue until the experimental section.

\section{Multi-View Treelet Transform}

To understand our generalization of the Treelet Transform to the multi-view setting, it is important to first make the connection between the Treelet Transform and matrix factorization, as is done in \cite{MMF, pMMF}.  Each step of the Treelet Transform attempts to reduce the objective $\textrm{off}(J^T \Sigma^{l-1} J)$ as much as possible, by zeroing the largest off-diagonal element of $\Sigma^{l-1}$.  Here, $l$ is the level of the hierarchical clustering and $\textrm{off}(\cdot)$ is the off-diagonal norm: $\textrm{off}(A) = \sum_{j \ne k} A_{j,k}^2$.  This is exactly the approach taken in the Jacobi diagonalization algorithm, where this iteration is taken to convergence.  In the multi-view setting, it is not possible to zero the largest off diagonal element of $\Sigma_i^{l-1}$, for all $1 \le i \le M$, with a single rotation $J$.  Instead,we find a $J$ that reduces a new objective, $\sum_i^M \textrm{off}(J^T \Sigma_i^{l-1} J)$, as much as possible, on each iteration.  This new objective is minimized in Joint Jacobi Diagonalization \cite{Cardoso} where the goal is to find a single basis consisting of a product of Jacobi rotations that diagonalizes a given collection of matrices.  We use this algorithm for computing joint rotations in MVTT.  \\
\vspace{0mm} \\
To compute $J$ in the multi-view setting, we must first find the two most similar columns across all views.  Thus we must compute

\[ (i, j, k) = \textrm{argmax}_{(i,j,k), j \in \{1,...,M\}} [ \rho_1[j,k], ..., \rho_M[j,k] ]. \]  

where $\Sigma_i$ is the covariance matrix of the $i$'th view and $\rho_i[j,k]$ is the correlation coefficient, $\frac{\Sigma_i[j,k]}{\sqrt{\Sigma_i[j,j]\Sigma_i[k,k]}}$.  Given the $j$'th and $k$'th columns, the question remains as to the choice of the rotation angle that will appropriately rotate them in each view.  The approach taken by the authors of \cite{Cardoso}, is to minimize $\sum_i^M \textrm{off}(J^T \Sigma_i^{l-1} J)$ by finding a simultaneous diagonalization of all $\Sigma^{l-1}_i$.  They note that this objective is minimized if we pick

\[ c = \sqrt{\frac{x+r}{2r}}, s = \sqrt{\frac{y-iz}{2r(x+r)}}, r = \sqrt{x^2+y^2+z^2}, \]  

where $[x,y,z]^T$ is an eigenvector associated with the largest eigenvalue of 

\[ G = \sum_i^M h(\Sigma^{l-1}_i)h(\Sigma^{l-1}_i)^T, \]

and $h$ is given by

\[ h(\Sigma^{l-1}_i) = [\Sigma^{l-1}_i[j,j]-\Sigma^{l-1}_i[k,k], \Sigma^{l-1}_i[j,k]+\Sigma^{l-1}_i[k,j]]. \]

The required rotation matrix $J$ is then equal to the identity except for the following submatrix:

\[
\left( \begin{array}{ccccccccc}
J[j, j] & J[j, k] \\
J[k, j] & J[k, k] \\
\end{array} \right) = 
\left( \begin{array}{ccccccccc}
c & -s \\
s & c \\
\end{array} \right).
\]

With the rotation matrices computed in this manner, we compute a Treelet Transform for multi-view data.  

\section{Theoretical properties}

Theoretical results are difficult to derive for this method.  The consistency arguments used by the authors of \cite{Treelet} in Theorem 1 rely on the fact that the Treelet Transform is a continuous function of the sample covariance matrix.  This is not the case for MVTT, as the eigenvectors of the sample covariance matrix are not continuous functions of it, thus the joint rotations computed in each step of MVTT are not continuous functions of the sample covariance.  As MVTT breaks the analytical notion of consistency as defined by the authors of \cite{Treelet}, it also breaks their analytical estimates of convergence.  We thus resort to numerical results to demonstrate theoretical properties such as convergence. \\
\vspace{0mm} \\
Note that in this section, when we discuss convergence, we mean convergence in the mean.  That is, for a sequence of random variables $\{X_n\}$ and the expectation operator, E, this sequence converges to $X$ in the mean if 

\[ \textrm{lim}_{n \rightarrow \infty} \textrm{E}(|X_n - X|) = 0. \]

\subsection{Synthetic data}

In this section, we construct synthetic datasets using the Kronecker Graph model.  The Kronecker Graph is a hierarchical graph with self-similar structure and has been used to model the time evolution of empirical graphs \cite{Kronecker}.  We generate a Kronecker Graph of depth three, $G_K$, by taking a 3x3 initiator matrix and raising it to the third Kronecker power.  The $k$'th Kronecker power of a matrix is the Kronecker product of this matrix with itself $k$ times.

Then, assuming Gaussian, additive noise, we compute new edge weights for $G_K$ as follows:

\[ w_n(j,k) = w_t(j,k) + \epsilon N(0,1), \]

where $w_t$ is the true edge weight, $w_n$ is the noisy edge weight, $N(0,1)$ is the Gaussian distribution of mean 0 and standard deviation 1, and $\epsilon$ is the noise level.  

\subsection{Convergence}

One of the motivations for this work was to find a method capable of discovering the structure of the true covariance matrix in the case where the size of our sample covariance matrices is fixed (e.g. our data is a graph with a fixed number of nodes).  In this case, we would like to show that the MVTT of the sample covariances is an increasingly good estimate of the Treelet Transform of the true covariance for large $M$.  More formally, where $T^l$ is the Treelet Transform operator for $L=l$, $MT^l$ is the MVTT operator for $L=l$, $\hat \Sigma_i$ is the $i$'th sample covariance matrix, and $\Sigma$ is the true covariance matrix, we measure the error as 

\[ E_M = ||\frac{1}{M}\left(\sum_i^M MT^l(\hat \Sigma_i)\right) - T^l(\Sigma)||_F^2. \]

Ideally, $E_M$ would converge to 0 as $M \rightarrow \infty$.  This would imply that for large $M$, our method is consistent and that we are able to completely recover all information lost to noise with the MVTT.  As shown in Figure 2, MVTT is not capable of completely recovering the true Treelet Transform from noisy covariance matrices and as $M \rightarrow \infty$, $E_M$ converges to some small number.

\begin{figure}[h]
  \includegraphics[width=0.5\linewidth]{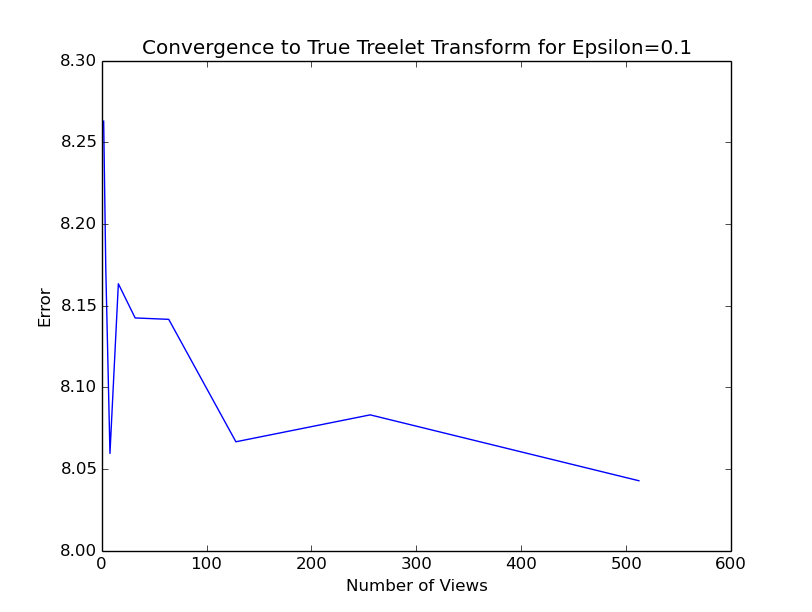}
  \includegraphics[width=0.5\linewidth]{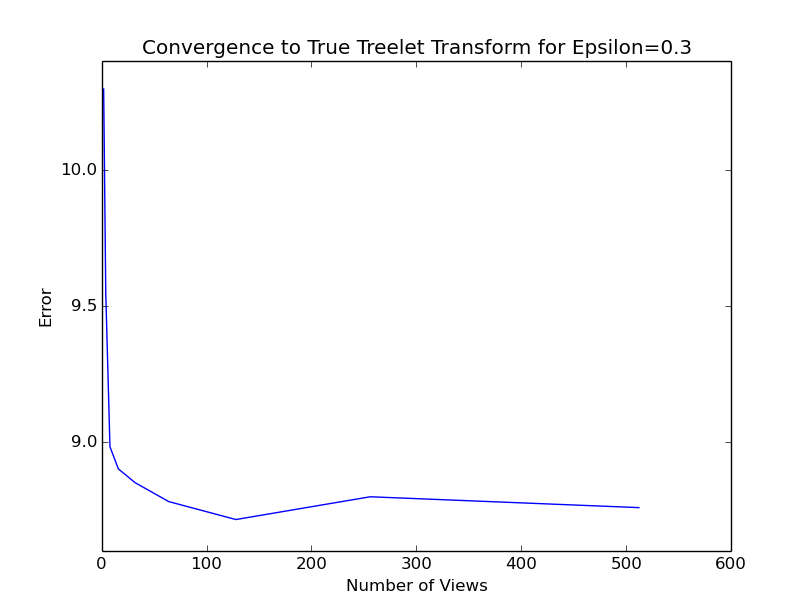}
  \includegraphics[width=0.5\linewidth]{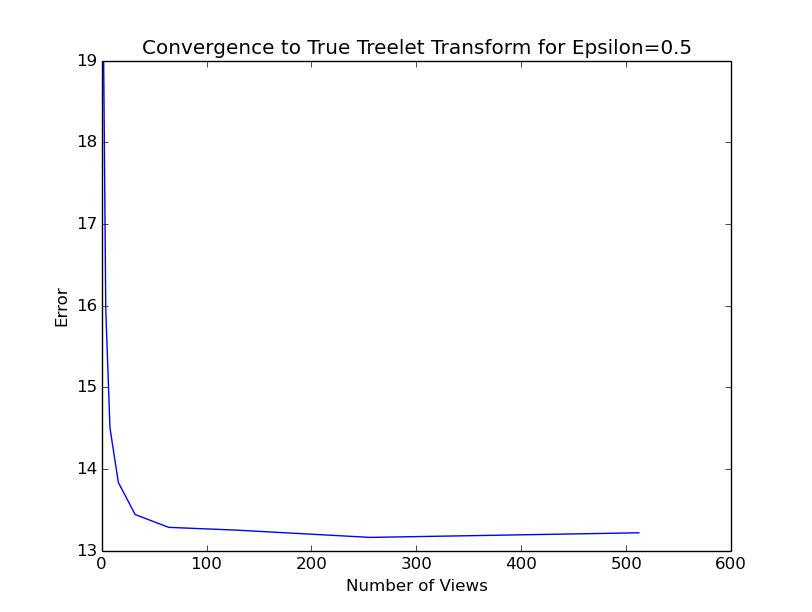}
  \includegraphics[width=0.5\linewidth]{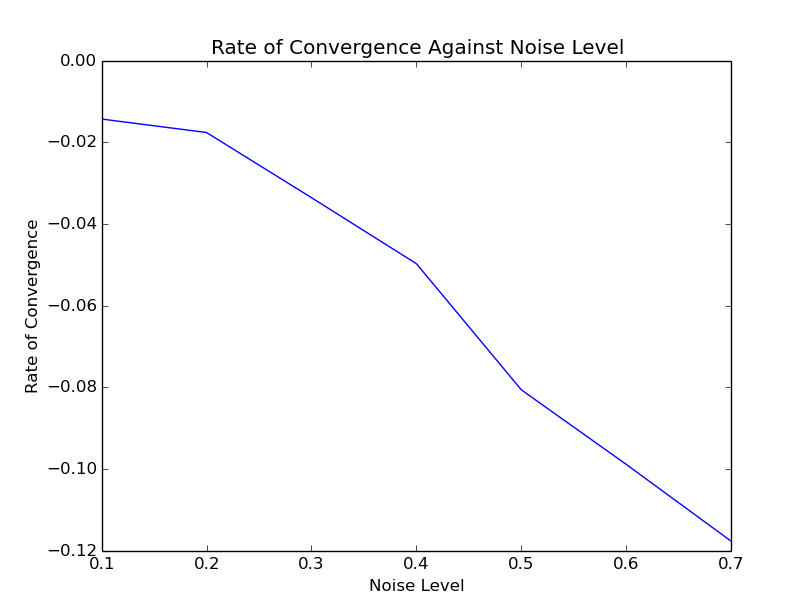}
  \caption{The first three plots demonstrate convergence of $E_M$ for various $M$ and $\epsilon$.  The final plot demonstrates that the rate of convergence of MVTT is approximately a linear function of the noise level.}
\end{figure}

To numerically demonstrate convergence of $E_M$, we performed bootstrap resampling.  For a fixed noise level, $\epsilon_B$, and a fixed number of views, $M_B$, we generated 20 collections of $M_B$ graphs, each with noise level $\epsilon_B$.  For each collection, we computed $E_M$, then averaged this error across all collections for a fixed $\epsilon_B$ and $M_B$.  The results of this test are given in Figure 2.  Of note is the effect of the value of $M$ on the value of $E_M$ for a given $\epsilon$.  Increasing values of $M$ have a greater effect on reducing the error for higher noise levels.  Thus one of the strengths of MVTT as opposed to the single view Treelet Transform is in finding hierarchical structure in data with high noise levels.  

As $E_M$ appears to be an exponential function of $M$, we also estimated the rate of convergence as the rate of decay of this exponential function.  These results are also given in Figure 2 and expanded upon in the Supplement.  Here we will simply point out that the rate of convergence is a decreasing function of the noise level, showing that MVTT converges faster for higher noise levels.  This property may be quite valuable in cases where the noise level is high, but additional views are difficult to obtain.

\section{Experiments}
\subsection{Denoising via basis coefficient thresholding}

To illustrate the ability of MVTT to capture consensus structure, we first show how it can be used to remove noise from data.  As in wavelet signal processing, we can remove certain kinds of noise from our data by thresholding insignificant basis coefficients.  Given the following representation of each $N$ element data point, $X_j$, in the $N\textrm{x}N$ matrix $X$:

\[ X_j = \sum_k^N c_k \tau_k, \]

where $\tau_k$ is the $k$'th basis vector and $c_k$ is the $k$'th basis coefficient.  The denoising problem can be defined as follows: we want to find some threshold, $T$, such that $||X_t-X_d||_F^2$, where $||\cdot||_F$ is the Frobenius norm, $X_t$ is the true data matrix, and $X_d$ is the denoised data matrix, is as small as possible.  In the synthetic case, we can compute this value directly, and so we can directly optimize it in computing $T$.  But in the experimental setting, we likely will not know the value of $X_t$, so finding $T$ becomes more complex.  \\
\vspace{0mm} \\
Similar to the work in \cite{FDR}, we use a hypothesis test to assign a p-value to each coefficient. Then, accepting a False Discovery Rate (FDR) as a parameter, we find a threshold for the basis coefficients.  Specifically, consider the following two-tailed hypothesis test:

H$_0$: $c_k=0$

H$_A$: $c_k \ne 0$

Assuming all $c_k$ are drawn from a Gaussian distribution centered at 0 with variance given by the variance of all $c_k$, we then have a method for directly computing a p-value, $p_k$, for each $c_k$.  Sorting the collection of all $p_k$, the FDR $p_{f}$ is the largest $p_k$ for which the null hypothesis is not rejected.  We then use the corresponding $c_f$ as a hard threshold for the basis coefficients.

\subsubsection{Synthetic data}

Using the noisy Kronecker Graph model described above, we test our thresholding approach on noise levels in $[0, 0.5]$.  In these results, we provide a qualitative comparison with the multi-view factorization method 

\[ \textrm{min}_{W_i, 1 \le i \le M; S} \sum_i^M||X_i-W_iS||_F \]

\[ W_i^TW_i = I, 1 \le i \le M, \]

which we refer to as non-probabilistic SRM, keeping with the terminology used in \cite{SRM}. We refer to the probabilistic solution of this optimization problem, also given in \cite{SRM}, simply as SRM.  We intend for this comparison to provide additional insight into the relative strengths of each method prior to our quantitative comparison later in Section 6.2.  We also sought to compare single-view with multi-view denoising.  To do so, we generated 100 views of a noisy Kronecker graph, computed the error of MVTT and the single view Treelet Transform, averaged both over all views, took the difference, and averaged this value over 30 trials.  Results from both experiments are displayed in Figure 3.  Of note is the fact that neither the probabilistic nor the non-probabilistic versions of SRM are able to capture the structure of the Kronecker Graph, where MVTT can.

\begin{figure}[h]
  \includegraphics[width=1.0\linewidth]{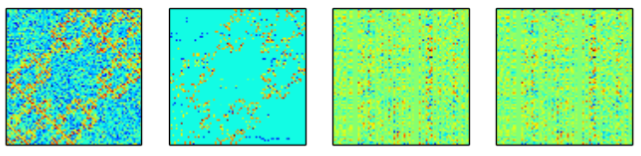}
  \centering
  \includegraphics[width=0.5\linewidth]{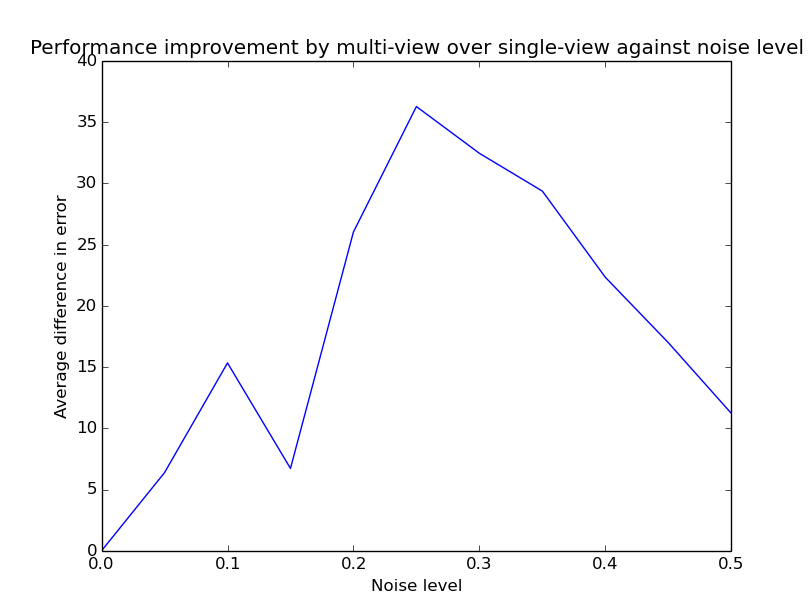}
  \caption{From the left, we have the original noisy image, the image denoised with MVTT, the image denoised with SRM, and the image denoised with non-probabilistic SRM.  The plot is of the average error of the multi-view approach subtracted from the average error of the single-view approach; the mean of 30 trials of this experiment are plotted.}
\end{figure}  

\subsubsection{Empirical data}

To further demonstrate the utility of MVTT for denoising, we tested it on a functional brain network derived from fMRI.  A functional brain network can be constructed by averaging voxel-level activations over larger brain regions and computing a correlation network between region-level time series.  Such networks have been proven useful in identifying a number of scientifically and potentially clinically significant phenomena.  For example, Schizophrenia has been identified as a disease characterized by an altered functional network topology \cite{Vertes}.  Further, it has been shown \cite{Bassett} that human functional brain networks have a hierarchical structure.  As these networks are also usually collected in replicate, they are an excellent choice for a dataset on which to test the denoising ability of MVTT.  Using the functional networks of 20 healthy subjects at rest \cite{Vertes}, we compared the output of MVTT with that obtained using domain specific assumptions.  Namely, these authors assume that the network should be connected and have a specific connection density.  Our results are shown in Figure 4.  

\begin{figure}[h]
  \includegraphics[width=1.0\linewidth]{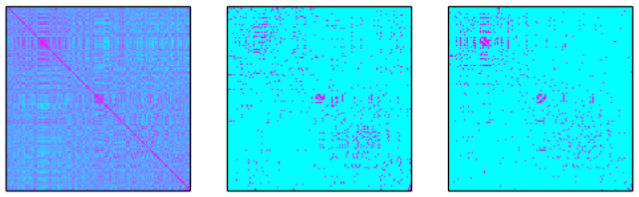}
  \caption{Denoised empirical functional networks.  The leftmost image is the empirical functional network, the middle image is the denoised version produced by MVTT, and the rightmost image is the output of the method given in \cite{Vertes}.}
\end{figure}  

Figure 4 demonstrates the similarity between the output of the denoising algorithm of \cite{Vertes} and that produced by MVTT.  With an FDR of 0.015, we find the results given in Table 1.  The authors of \cite{Vertes} consider a connection density between 0.04 and 0.16 physiologically reasonable.  MVTT produces a denoised graph that approximates their assumptions while being highly correlated with their output.  MVTT is thus able to capture the structure of the empirical functional networks with a more general set of assumptions.

\begin{table}[h]
  \caption{MVTT Denoising of Functional Networks from \cite{Vertes}}
  \label{sample-table}
  \centering
  \begin{tabular}{lll}
    \toprule
    \cmidrule{1-2}
    Measurement     & Value     & Standard Deviation  \\
    \midrule
    Connection Density & 0.147  & 0.066     \\
    Number of Connected Components     & 1.65 & 0.853      \\
    Pearson Correlation     & 0.462       & 0.037  \\
    \bottomrule
  \end{tabular}
\end{table}

\subsection{Shared response}

The Shared Response Problem (SRP) is formulated in \cite{SRM} as one of computing correlations between voxel-level fMRI timeseries.  Because of their scientific and clinical potential, we reformulate the SRP based on correlations between functional networks.  Given the hierarchical structure of functional networks and the ability of MVTT to capture this structure, MVTT would seem an excellent tool for finding consensus structure across replicates which may have a shared response.  \\
\vspace{0mm} \\
To compute the shared response, we use the \textit{Raider} dataset first presented in \cite{Haxby}.  Our method for computing the shared response is similar to that presented by the authors in \cite{SRM}.  For each hemisphere, we split the dataset into a training half and testing half based on time.  Both the training and testing sets are then split into two groups of subjects, $G_1$ and $G_2$.  MVTT and SRM are both run on the training halves of $G_1$ and $G_2$, resulting in bases $Q_1$ and $Q_2$ for MVTT and $W^1_i$ and $W^2_i$ for SRM where $1 \le i \le M$.  $Q_1$ and $Q_2$ and the coefficient matrices returned by SRM, $S_1$ and $S_2$, are then registered by solving orthogonal Procrustes problems.  Denoising is performed then the Pearson Correlation is computed between groups of testing subjects as a measure of the shared response between these two groups.  The averages over five different subject partitions are given in Table 2. 

\begin{table}[h]
  \caption{Shared Response of \textit{Raider} Dataset}
  \label{sample-table}
  \centering
  \begin{tabular}{lllll}
    \toprule
    \cmidrule{1-2}
    Method & LH-Corr & LH-STD & RH-Corr & RH-STD  \\
    \midrule
    None & 0.363 & 0.006 & 0.542 & 0.005 \\
    MVTT-feature space & 0.691 & 0.002 & 0.705 & 0.002 \\
    MVTT-label space & 0.687 & 0.003 & 0.6994 & 0.001 \\
    SRM-feature space & 0.048 & 0.004 & 0.088 & 0.009 \\
    SRM-label space & 0.095 & 0.003 & 0.173 & 0.004 \\
    \bottomrule
  \end{tabular}
\end{table}

The results in Table 2 are consistent with the experiments shown in Section 6.1.1: SRM is unable to capture the hierarchical structure of the functional networks and information is lost in projecting the data onto the bases returned by it.  On the other hand, MVTT is successful in capturing the consensus structure in the hierarchical graphs and as such, the computed shared response using MVTT is higher.  The feature and label space results are given after denoising with an FDR of 0.01 and it should be noted that the denoising process resulted in a reduction in the performance of SRM, with original feature space correlations of 0.324 ($\sigma = 0.007$) and 0.201 ($\sigma = 0.003$) for the right and left hemispheres respectively.

\section{Conclusion}

We have introduced a novel method, the Multi-View Treelet Transform, and shown numerically that it has desirable convergence properties.  Moreover, we have shown how it may be applied to three problems to achieve state-of-the-art results: the denoising of synthetic, hierarchical graphs, the denoising of empirical functional networks, and a reformulation of the Shared Response Problem.

\pagebreak
\title{Supplemental material}

\makesimpletitle




\section{Theoretical properties}
\subsection{Synthetic data}

To provide additional motivation and intuition for why MVTT and the Treelet Transform should capture the structure of a Kronecker graph, consider the following simple hierarchical clustering procedure.  

1.  Break the adjacency matrix of $G_K$, $A_{G_K}$, up into contiguous, non-overlapping blocks of 3x3 matrices.

2.  Apply a normalized box filter to each block, $B$, call the returned average $A_B$.

3.  Compress all entries of $B$ into a single entry with value $A_B$.

4.  Iterate over this approach.

If this procedure is performed repeatedly, we obtain the results shown in Figure 1.  Thus we can see that hierarchical clustering via iterative smoothing can capture self-similar, hierarchical structure in graphs.  

\begin{figure}[h]
  \includegraphics[width=0.5\linewidth]{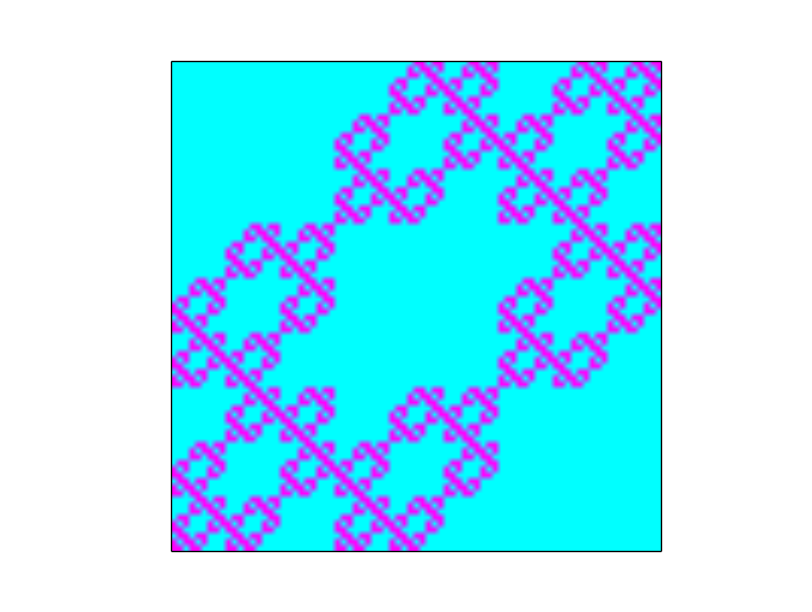}
  \includegraphics[width=0.5\linewidth]{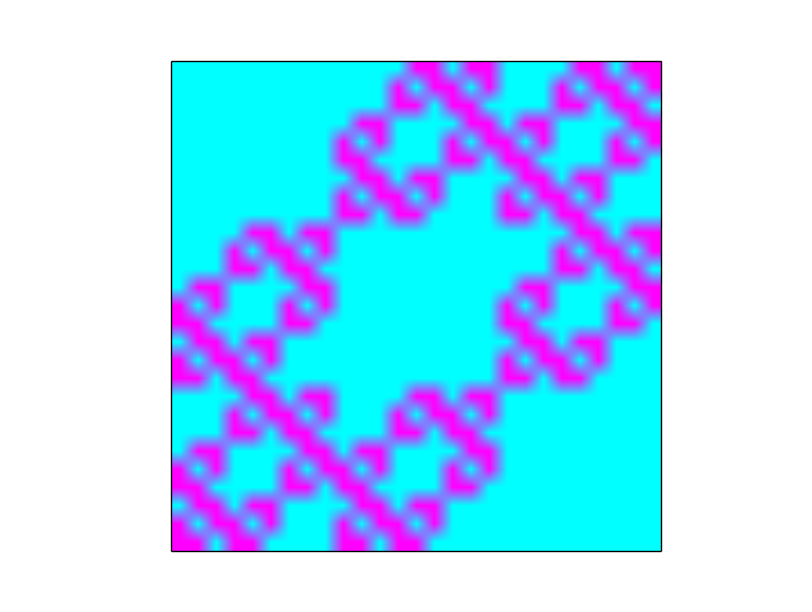}
  \includegraphics[width=0.5\linewidth]{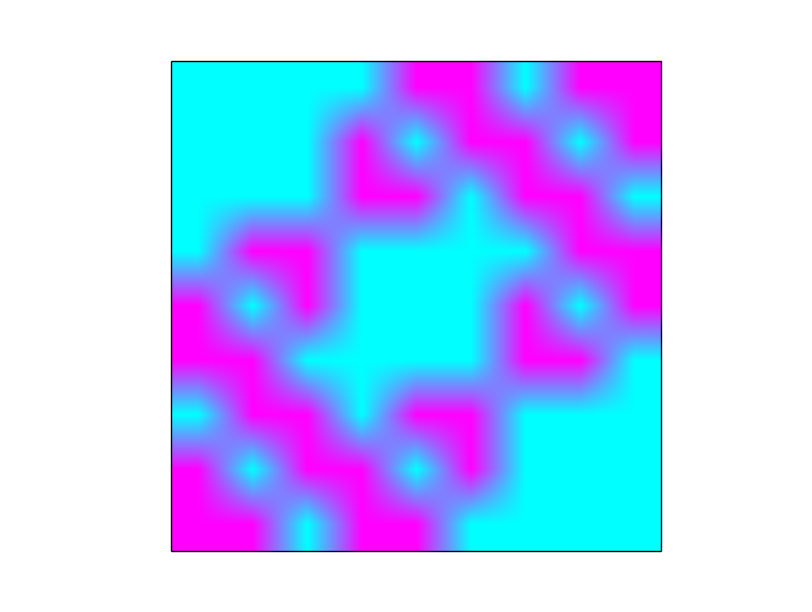}
  \includegraphics[width=0.5\linewidth]{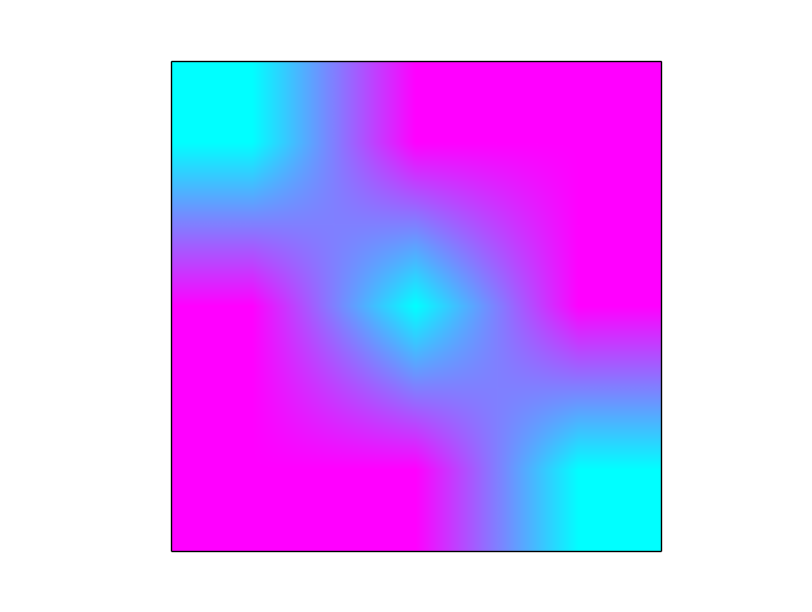}
  \caption{Iterative smoothing is capable of capturing self-similar structure in this Kronecker Graph.}
\end{figure}

\subsection{Stability}

To further explore the theoretical properties of MVTT, we provide an analysis of its stability below.

\begin{table}[h]
  \caption{Stability of MVTT}
  \label{sample-table}
  \centering
  \begin{tabular}{lll}
    \toprule
    \cmidrule{1-2}
    Noise Level     & Stability     & Standard Deviation  \\
    \midrule
    0.1 & 0.248  & 0.027     \\
    0.2 & 0.239 & 0.063      \\
    0.3 & 0.248 & 0.039  \\
    0.4 & 0.214 & 0.054  \\
    0.5 & 0.340 & 0.206  \\
    \bottomrule
  \end{tabular}
\end{table}

For two distinct collections of $M$ views, $C_1$ and $C_2$, we would like for $C_1$ and $C_2$ to be reasonably ``close''.  We can define this "closeness" as stability if we then define stability as the standard deviation over all $E_M$, for a given noise level and value of $M$.  $C_1$ and $C_2$ are confirmed to be close in Table 1, with one exception for $\epsilon = 0.5$.  The reduction in stability for this noise level arises from variable performance for low values of $M$ (shown in Figure 2), a result which is consistent with our findings for convergence and denoising.  Specifically, the Treelet Transform for high levels of noise returns a basis that increasingly reflects the structure of this noise and as such, for low values of $M$, the output of MVTT is less stable.

\subsection{Rate of convergence}

Another finding concerning the convergence of MVTT is that $E_M$ appears to be an exponentially decaying function of $M$, for example, of the form:

\[ E_M = e^{rM} + \textrm{Bias}(\hat \Theta, \Theta), \]

where $\hat \Theta$ is the estimator given by the average of $MT^l$ over all $M$ views and $\Theta$ is $T^l(\Sigma)$.  $\textrm{Bias}(\hat \Theta, \Theta)$ is then the value of $E_M>0$ to which the error estimates converge.  This observation is confirmed in Figure 2 of the main text, where it is shown that the rate of convergence of MVTT is a decreasing linear function of the noise level.  And since the rate of convergence is 0 when $\epsilon=0$, for some $s<0$, we have the following relationship for the error:

\[ E_M = e^{s\epsilon M} + \textrm{Bias}(\hat \Theta, \Theta) \]

\section{Experiments}
\subsection{Denoising of synthetic data}

To provide a more complete qualitative picture of the performance of MVTT relative to SRM and non-probabilistic SRM in denoising Kronecker graphs, we give additional examples below.  

\begin{figure}[h]
  \includegraphics[width=1.0\linewidth]{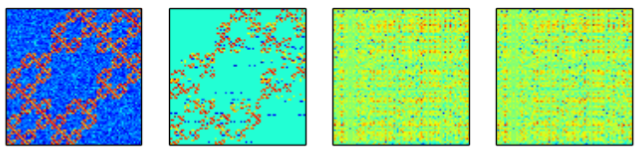}
  \includegraphics[width=1.0\linewidth]{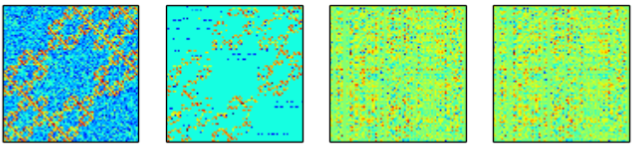}
  \includegraphics[width=1.0\linewidth]{/home/bam/Documents/submission/figures/mvtt_v_srm_denoise_03_ss.png}
  \includegraphics[width=1.0\linewidth]{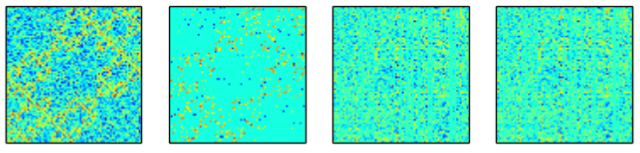}
  \includegraphics[width=1.0\linewidth]{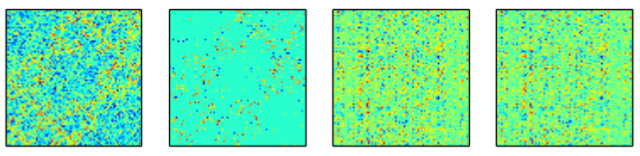}
  \caption{Noise levels of graphs are given top to bottom: 0.1, 0.2, 0.3, 0.4, 0.5.  Left to right, the images are of the original noisy Kronecker graph, the graph denoised with MVTT, the graph denoised with SRM, and the graph denoised with non-probabilistic SRM.}
\end{figure}

\subsection{Shared response computation}

The number of basis vectors computed by SRM was set to the maximum possible for both Kronecker graph denoising as well as shared response computation.  The motivation for this is given in Figures 3 and 4.  Of note is the fact that even assuming that the Kronecker graph is full rank, there is significant distortion in the reconstruction by the probabilistic variant of SRM.  

\begin{figure}[h]
  \includegraphics[width=0.95\linewidth]{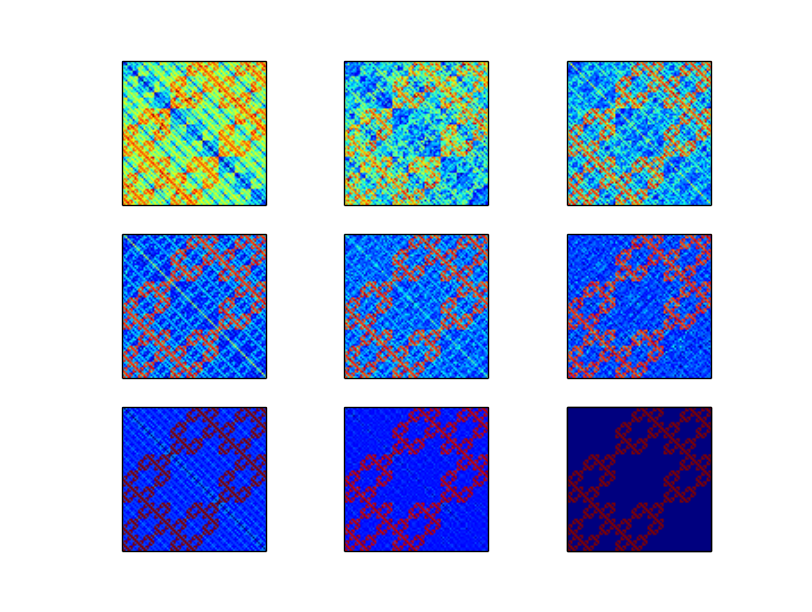}
  \caption{Reconstruction of Kronecker graph by non-probabilistic SRM for various numbers of basis vectors.  Top to bottom, left to right: 9, 18, 27, 36, 45, 54, 63, 72, and 81 (full rank).}
  \includegraphics[width=0.95\linewidth]{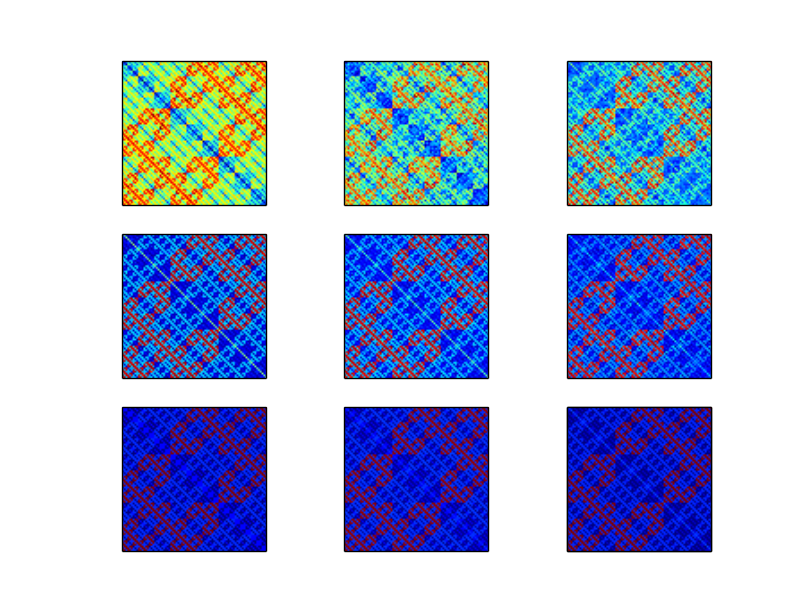}
  \caption{Reconstruction of Kronecker graph by probabilistic SRM for various numbers of basis vectors.  Top to bottom, left to right: 9, 18, 27, 36, 45, 54, 63, 72, and 81 (full rank).}
\end{figure}


\clearpage
\begin{thebibliography}{9}

\bibitem{Bassett}

Bassett, D.S., Bullmore, E., Verchinski, B.A., Mattay, V.S., Weinberger, D.R., and Lindenberg, A.M.  (2008).  Hierarchical organization of human cortical networks in health and schizophrenia.  \textit{The Journal of Neuroscience} \textbf{28}(37): 9239-9248. 

\bibitem{BLM}

Tenenbaum, J.B. and Freeman, W.T.  (2000).  Seperating style and content with bilinear models.  \textit{Neural Computation} \textbf{12}(6): 1247-1283.  2000.

\bibitem{Cardoso}

Cardoso, J.F. and Souloumiac, A.  (1996).  Jacobi angles for simultaneous diagonalization.  \textit{SIAM Journal on Matrix Analysis and Applications} \textbf{17}(1): 161-164.

\bibitem{CCA}

Hardoon, D.R., Szedmak, S., and Shawe-Taylor, J.  (2004).  Canonical correlation analysis; An overview with application to learning methods.  \textit{Neural Computation} \textbf{16}(12): 2639-2664.

\bibitem{CoRegSpecClust}

Kumar, A., Rai, P., and Daume, H.  (2011).  Co-regularized multi-view spectral clustering.  \textit{Advances in Neural Information Processing Systems 24}: 1413-1421.

\bibitem{DW}

Coifman, R.R. and Maggioni, M.  (2006).  Diffusion wavelets.  \textit{Applied and Computational Harmonic Analysis} \textbf{21}(1): 53-94.

\bibitem{equiNMF}

Hidru, D. and Goldenberg, A.  (2014).  EquiNMF: graph regularized multiview nonnegative matrix factorization.  \textit{arXiv}:1409.4018.

\bibitem{FDR}

Abramovich, F. and Benjamini, Y.  (1996).  Adaptive thresholding of wavelet coefficients.  \textit{Computational Statistics and Data Analysis} \textbf{22}(4): 351-361.

\bibitem{GMA}

Sharma, A., Kumar, A., and Daume, H.  (2012).  Generalized multiview analysis: a discriminative latent space.  \textit{Computer Vision and Pattern Recognition}: 2160-2167.

\bibitem{Haxby}

Haxby, J.V., Guntupalli, J.S., Connolly, A.C., Halchenko, Y.O., Conroy, B.R., Gobbini, M.I., Hanke, M., and Ramadge, P.J.  (2011).  A common, high-dimensional model of the representational space in human ventral temporal cortex.  \textit{Neuron} \textbf{72}(2): 404-416.

\bibitem{Kronecker}

Leskovec, J., Chakrabarti, D., Kleinberg, J., Faloutsos, C., and Ghahramani, Z.  (2010).  Kronecker graphs: an approach to modeling networks.  \textit{Journal of Machine Learning Research} \textbf{11}(3): 985-1042.

\bibitem{MMF}

Kondor, R., Teneva, N., Garg, V.  Multiresolution matrix factorization.  \textit{International Conference on Machine Learning 31}: 1620-1628.

\bibitem{multiNMF}

Liu, J., Wang, C., Gao, J., and Han, J.  (2013).  Multi-view clustering via joint non-negative matrix factorization.  \textit{Proceddings of the SIAM Conference on Data Mining}.

\bibitem{multiFactorNMF}

Lyu, S. and Wang, X.  (2013).  On algorithms for sparse multi-factor NMF.  \textit{Advances in Neural Information Processing Systems 26}: 602-610.  

\bibitem{pMMF}

Kondor, R., Teneva, N., and Mudrakarta, P.K.  (2015).  Parallel MMF: a multiresolution approach to matrix computation.  \textit{arXiv}:1507.04396.

\bibitem{PLS}

Abdi, H.  (2007).  Partial least squares regression.  \textit{Encyclopedia for research methods for the social sciences}: 792-795.

\bibitem{SRM}

Chen, P.H., Chen, J., Yeshurun, Y., Hasson, U., Haxby, J.V., and Ramadge, P.J.  (2015).  A reduced-dimension fMRI shared response model.  \textit{Advances in Neural Information Processing Systems 28}: 460-468.

\bibitem{symNMF}

Kuang, D., Ding, C., and Park, H.  (2012).  Symmetric nonnegative matrix factorization for graph clustering.  \textit{Proceedings of the SIAM Conference on Data Mining}.

\bibitem{Tang}

Tang, W., Lu, Z., and Dhillon, I.S.  (2009).  Clustering with multiple graphs.  \textit{Proceedings of the International Conference on Data Mining 9}: 1016-1021.

\bibitem{Treelet}

Lee, A.B., Nadler, B., and Wasserman, L.  (2008).  Treelets-an adaptive multi-scale basis for sparse unordered data.  \textit{The Annals of Applied Statistics} \textbf{2}(2): 435-471.

\bibitem{Vertes}

Vertes, P.E., Bloch-Alexander, A.F., Gogtay, N., Giedd, J.N., Rapoport, J.L., and Bullmore, E.T.  (2011).  Simple models of human brain functional networks.  \textit{Proceedings of the National Academy of Sciences} \textbf{109}(15): 5868-5873.

\end{thebibliography}
\end{document}